%% file: acl_latex.tex
\pdfoutput=1

\documentclass[11pt]{article}

\usepackage[final]{acl}

\usepackage{times}
\usepackage{latexsym}

\usepackage[T1]{fontenc}

\usepackage[utf8]{inputenc}

\usepackage{microtype}

\usepackage{inconsolata}

\usepackage{graphicx}

\usepackage{longtable}
\usepackage{xcolor}
\usepackage{colortbl}
\usepackage{multicol}
\usepackage{booktabs}
\usepackage{makecell}
\usepackage{amsmath, amssymb}
\usepackage{multirow}
\usepackage{mathtools}
\usepackage[inline]{enumitem}
\usepackage{subcaption}
\usepackage{float}
\usepackage{graphicx}
\usepackage{caption} 
\usepackage{upgreek}
\usepackage{seqsplit}
\usepackage{color,soul}
\usepackage{arydshln}
\usepackage{tabularx}
\usepackage{adjustbox}
\usepackage{array}
\usepackage{algorithm, algpseudocode}
\usepackage{setspace}

\input{math_commands}

\usepackage{caption}
\usepackage{subcaption}
\title{Knowledge Base Construction for Knowledge-Augmented Text-to-SQL}

\author{
    \textbf{Jinheon Baek}$^1$\thanks{Work done during an internship at IBM Research.} \; 
    \textbf{Horst Samulowitz}$^2$ \; 
    \textbf{Oktie Hassanzadeh}$^2$ \\
    \textbf{Dharmashankar Subramanian}$^2$ \;
    \textbf{Sola Shirai}$^2$ \;
    \textbf{Alfio Gliozzo}$^2$ \;
    \textbf{Debarun Bhattacharjya}$^2$ \\
    KAIST$^1$ \;\; IBM Research$^2$ \\
    {
        \texttt{jinheon.baek@kaist.ac.kr} \; \texttt{\{samulowitz, hassanzadeh\}@us.ibm.com}
    } \\
    {
        \texttt{dharmash@us.ibm.com} \; \texttt{solashirai@ibm.com} \; \texttt{\{gliozzo, debarunb\}@us.ibm.com}
    }
}

\begin{document}
\maketitle

\input{Sections/1_abstract}
\input{Sections/2_introduction}

\input{Sections/3_related_work}
\input{Sections/4_method}

\input{Sections/5_experimental_setup}
\input{Sections/6_experimental_result}
\input{Sections/7_conclusion}

\bibliography{custom}

\input{Sections/8_appendix}

\end{document}

%% file: math_commands.tex
\usepackage{amsmath,amsfonts,bm}

\def\eqref#1{equation~\ref{#1}}

\def\1{\bm{1}}

\def\vk{{\bm{k}}}

\def\vq{{\bm{q}}}

\def\vs{{\bm{s}}}

\def\mD{{\bm{D}}}

\DeclareMathAlphabet{\mathsfit}{\encodingdefault}{\sfdefault}{m}{sl}
\SetMathAlphabet{\mathsfit}{bold}{\encodingdefault}{\sfdefault}{bx}{n}



%% file: Sections/1_abstract.tex
\begin{abstract}

Text-to-SQL aims to translate natural language queries into SQL statements, which is practical as it enables anyone to easily retrieve the desired information from databases. Recently, many existing approaches tackle this problem with Large Language Models (LLMs), leveraging their strong capability in understanding user queries and generating corresponding SQL code. Yet, the parametric knowledge in LLMs might be limited to covering all the diverse and domain-specific queries that require grounding in various database schemas, which makes generated SQLs less accurate oftentimes. To tackle this, we propose constructing the knowledge base for text-to-SQL, a foundational source of knowledge, from which we retrieve and generate the necessary knowledge for given queries. In particular, unlike existing approaches that either manually annotate knowledge or generate only a few pieces of knowledge for each query, our knowledge base is comprehensive, which is constructed based on a combination of all the available questions and their associated database schemas along with their relevant knowledge, and can be reused for unseen databases from different datasets and domains. We validate our approach on multiple text-to-SQL datasets, considering both the overlapping and non-overlapping database scenarios, where it outperforms relevant baselines substantially.

\end{abstract}

%% file: Sections/2_introduction.tex
\section{Introduction}

Text-to-SQL aims to transform natural language queries from users into Structured Query Language (SQL) statements, to interact with and retrieve the information from databases~\cite{text2sql/1, text2sql/2, text2sql/3, text2sql/4}, as illustrated in Figure~\ref{fig:concept} (A). This task has recently gained much attention since it allows non-experts to access and manipulate database information without needing to understand complex database languages. In the meantime, Large Language Models (LLMs) have shown impressive capabilities in processing and generating text and code, which have been further extended for text-to-SQL~\cite{LLM/Text2SQL/1, LLM/Text2SQL/2}.

Despite their huge successes, transforming user queries into SQL statements may still be challenging due to the need for specific domain knowledge and an understanding of the underlying database schemas, which poses a significant hurdle even for the most advanced LLMs to achieve high accuracy across diverse datasets~\cite{bird}. For example, consider a scenario where the user asks for the query: "What is the WACC for Company X?". To accurately translate this into an SQL statement, the text-to-SQL model should understand the concept and calculation of Weighted Average Cost of Capital (WACC), which involves multiple factors including the cost of equity, cost of debt, and the respective proportions of each in the capital structure. In addition, the model needs to comprehend the specific schema of the financial database, where relevant data is distributed across multiple tables such as 'Equity', 'Debt', and 'Capital Structure'. 

\input{Figures/concept}

To tackle the aforementioned limitations due to the lack of the domain-specific knowledge for SQL generation, recent studies have proposed collecting and annotating explicit knowledge, which is then leveraged for SQL generation~\cite{knowledge-intensive-text-to-sql, bird}. However, while these approaches substantially improve the performance of existing text-to-SQL models, they rely on extensive human annotations, which may be suboptimal (and nearly impractical) to conduct for all queries considering a diverse source of domain-specific knowledge from numerous databases. To address this issue, recent work proposes generating a few pieces of knowledge for each query based on the query itself and its relevant database schema~\cite{Knowledge-to-SQL} (see Figure~\ref{fig:concept} (B)). However, although this method demonstrates promise in automatic knowledge generation, certain knowledge required for one query can be directly reused or provide insights for multiple queries within the same database, as shown in Figure~\ref{fig:concept} (Right). Also, this knowledge can be generalizable to other queries for different databases. 

Motivated by these observations, this work proposes an automatic approach to build a knowledge base, designed to serve as a comprehensive repository of domain-specific knowledge for text-to-SQL and capable of providing knowledge for multiple queries with the same database and even across the different databases. To construct this knowledge base, we generate knowledge entries based on available samples and their associated database schemas through LLM prompting, and then compile all of them together. During this prompting process, we provide LLMs with relevant examples to contextualize and guide the generation of useful knowledge in the right format that is further grounded in the database schema. Then, once constructed, the knowledge base allows for the retrieval of relevant knowledge for the given test-time query, which is then used alongside the query to formulate the SQL statement. Note that while ideally the knowledge base would cover all possible queries, it may not always do so. Nevertheless, the existing knowledge in it could still offer valuable insights for generating the required knowledge for new queries. Thus, by leveraging similar knowledge from the knowledge base, we further prompt LLMs to produce the most suitable knowledge for the query at inference time. We call our method Knowledge-Augmented Text-to-SQL (KAT-SQL), depicted in Figure~\ref{fig:concept} (C).

We experimentally validate the proposed KAT-SQL on two different text-to-SQL scenarios, involving both the overlapping and non-overlapping databases between training and test phases, showing that the proposed knowledge base construction-based text-to-SQL approach surpasses the existing (knowledge-augmented) text-to-SQL baselines. We also assess the generalizability of our knowledge base constructed from one dataset by applying it to different datasets that lack any annotated knowledge, demonstrating that our knowledge base is versatile and can effectively improve SQL generation for even unseen databases from other datasets.

%% file: Figures/concept.tex
\begin{figure*}[t!]
    \begin{minipage}{0.85\textwidth}
    \centering
    \includegraphics[width=1.0\linewidth]{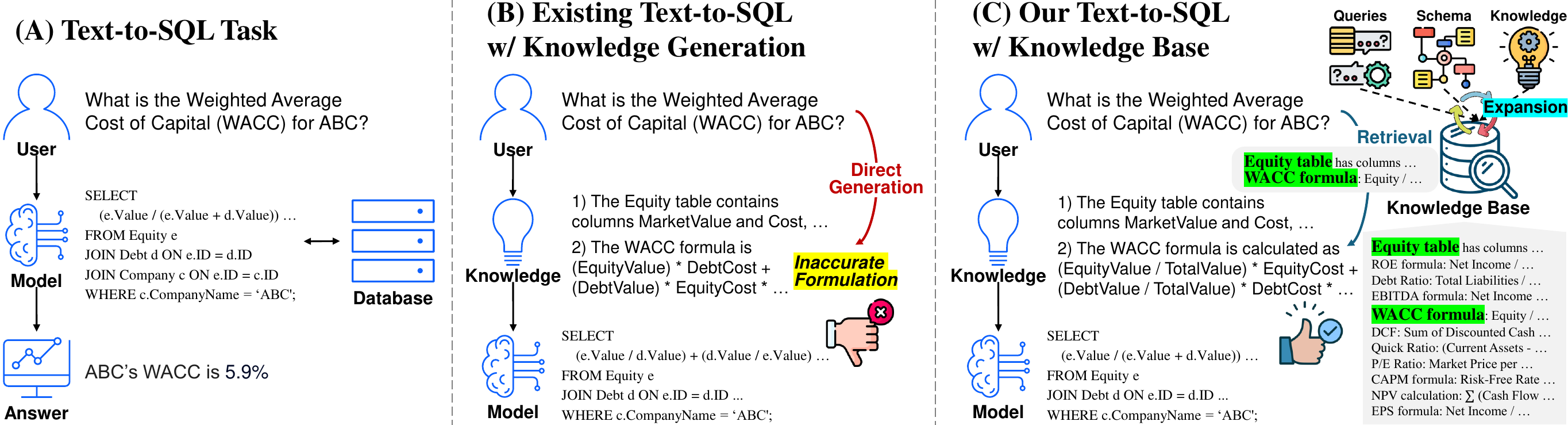}
    \end{minipage}
    \hfill
    \begin{minipage}{0.14\textwidth}
    \includegraphics[width=1.0\linewidth]{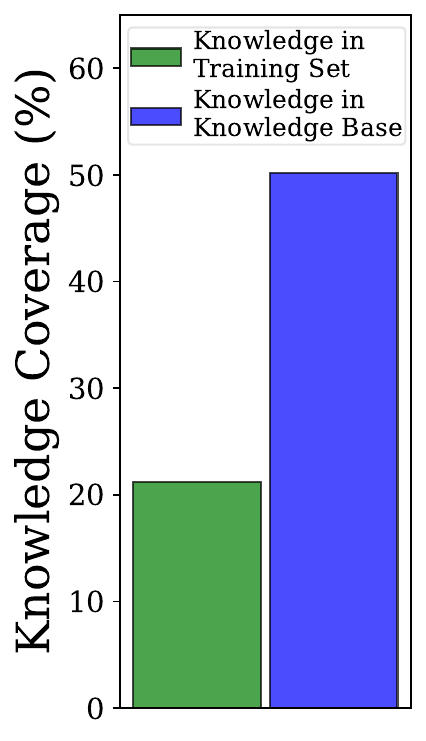}
    \end{minipage}
    \vspace{-0.125in}
    \caption{(A) Text-to-SQL aims to translate a user query into a SQL statement executable over a database, to access the desired information. (B) Existing Text-to-SQL with Knowledge Generation approaches first generate the knowledge relevant to the user query and then formulate the SQL statement with this generated knowledge. (C) Our Text-to-SQL with Knowledge Base Construction approach builds the repository of the knowledge and then reuses the knowledge within it across multiple queries and databases. (Right:) We observe that the knowledge in the training set of the text-to-SQL benchmark dataset~\cite{bird} covers 21\% of the knowledge required for test-time queries, and our constructed knowledge base further covers 50\% of them.}
    \label{fig:concept}
    \vspace{-0.1in}
\end{figure*}

%% file: Sections/3_related_work.tex
\vspace{-0.025in}
\section{Related Work}

\vspace{-0.025in}
\paragraph{LLM-Powered Text-to-SQL}
LLMs have shown remarkable performances across a wide range of tasks~\cite{GPT-4, Gemini, llama3modelcard}, including text-to-SQL, due to their strong capability in understanding natural language and generating structured code~\cite{LLM/Text2SQL/1, LLM/Text2SQL/2}. Specifically, various studies have developed and advanced the prompting techniques for text-to-SQL, for example, using Chain-of-Thought (CoT)~\cite{CoT, text2sql/cot/1, text2sql/cot/2}, investigating sophisticated prompt design strategies~\cite{text2sql/prompt/1}, and aggregating LLM-generated outputs from multiple prompts~\cite{text2sql/consistent/1, C3} akin to self-consistency~\cite{Self-Consistency}. In addition, another line of study proposes decomposing the text-to-SQL problem into multiple subtasks, and feeding the solutions of subtasks (from multiple models or agents) into the LLM to derive the final SQL statement~\cite{SC-Promp, DIN-SQL, MAC-SQL}. The knowledge internalized in LLMs might however not be sufficient to handle diverse queries, which oftentimes requires grounding in the database schemas or additional domain-specific information for specialized domains, which gives rise to the need for leveraging external knowledge for text-to-SQL.

\vspace{-0.025in}
\paragraph{Knowledge-Augmented Text-to-SQL}
There are a few recent studies that propose augmenting text-to-SQL models with explicit knowledge. Specifically, \citet{knowledge-intensive-text-to-sql} collect formulaic knowledge (e.g., Trade Balance = Exports – Imports) available from public resources such as finance reports and store the collected knowledge into a knowledge bank with proper human-involved post-processing. The text-to-SQL model then retrieves relevant knowledge for any given query from the knowledge bank and uses it to convert the query into the SQL statement. In addition, \citet{bird} release a large-scale benchmark dataset for the text-to-SQL task, where each question is associated with specific knowledge that is manually annotated by humans. Manual annotation is however costly and time consuming, requiring effort and expertise on the part of domain-experts. To address this challenge, more recent work proposes automatically generating the knowledge based on the question and database schema, and utilizing this knowledge for text-to-SQL~\cite{Knowledge-to-SQL}. In our work, instead of generating only a few pieces of knowledge for each question, we propose to construct a comprehensive knowledge base. This provides a repository of reusable knowledge that can be leveraged across multiple queries, which can be further adapted to various databases over different domains in a scalable way, in contrast to existing work.

\paragraph{Data Generation with LLMs}
The recent advent of LLMs has revolutionized the field of data generation, as they can produce vast amounts of high-quality samples without costly human annotation. Specifically, several efforts around LLM-based synthetic data generation, such as Self-Instruct~\cite{Self-Instruct}, Alpaca~\cite{alpaca}, Evol-Instruct~\cite{evol-instruct}, Orca~\cite{orca}, and InstructLab~\cite{instructlab}, propose generating a large number of samples from LLMs by prompting them. Also, motivated by the capabilities of LLMs in generating synthetic data and memorizing factual knowledge, some other work aims to populate an encyclopedic knowledge base like Wikidata~\cite{Wikidata} with LLMs~\cite{LLM/KBC/1, LLM/KBC/2, LLM/KBC/3}. Most of the knowledge in such encyclopedic knowledge bases is however unsuitable for text-to-SQL since it is neither relevant to formulate SQL statements from user queries nor aware of database schemas necessary for the query conversion. Thus, unlike them, our approach stands apart as the first to automatically construct a text-to-SQL knowledge base.

%% file: Sections/4_method.tex
\section{Method}

In this section, we present Knowledge-Augmented Text-to-SQL (KAT-SQL), an approach that automatically constructs a knowledge base and utilizes the relevant knowledge from it for text-to-SQL.

\subsection{Problem Statement}

We begin with formally explaining text-to-SQL and the knowledge augmentation technique for it.

\paragraph{Text-to-SQL}
Text-to-SQL aims to translate a natural language query from a user into a syntactically correct and semantically precise SQL statement. Formally, let $\vq$ be the user query (consisting of a sequence of tokens) and $\mathcal{D}$ be the database schema containing multiple tables and columns. Then, the SQL generation model $f$ can be represented as follows: $\vs = f(\vq, \mathcal{D})$ where $\vs$ is the SQL statement (consisting of a sequence of tokens) that attempts to retrieve the information requested by $\vq$ over $\mathcal{D}$. 

In this work, we operationalize $f$ with LLMs, to harness their strong capability in understanding the semantics of $\vq$ and generating the corresponding SQL code $\vs$, as follows: $\vs = \texttt{LLM}_{\theta}(\mathcal{T}(\vq, \mathcal{D}))$ where $\theta$ is the model parameters and $\mathcal{T}$ is the prompt template. Typically, the model parameters $\theta$ remain fixed due to the high costs associated with further fine-tuning of them and sometimes their limited accessibility. Also, the prompt template $\mathcal{T}$ serves as a structured format that outlines the context, which includes task descriptions and instructions as well as few-shot demonstrations, to guide the model in generating accurate SQL codes.

Notably, while there have been great successes in advancing the LLM itself and optimizing its usage for text-to-SQL, such as using advanced prompting techniques or breaking down the task into multiple subtasks~\cite{CoT, text2sql/cot/1, text2sql/cot/2, SC-Promp, DIN-SQL, MAC-SQL}, these improvements alone may not be sufficient to fully handle queries that require the deep domain knowledge or precise understanding of complex database schemas. In other words, the internal parametric knowledge of LLMs, while robust, may not fully encompass the diverse range of query variations and database structures, especially when these databases have distinct schemas or certain specialized terminology. 

\paragraph{Knowledge-Augmented Text-to-SQL}
To tackle the aforementioned limitations, we focus on augmenting text-to-SQL with the knowledge relevant to the query, providing valuable insights into the domain-specific terminology and complex database schemas. If we denote this knowledge as $\vk$, then the previous text-to-SQL process is redefined to incorporate it, as follows: $\vs = \texttt{LLM}_{\theta}(\mathcal{T}(\vq, \vk, \mathcal{D}))$.

While there have been few studies that explore this knowledge-augmented text-to-SQL paradigm, there are still a couple of challenges. Specifically, \citet{knowledge-intensive-text-to-sql} and \citet{bird} propose collecting and annotating the explicit knowledge required to convert queries into SQL statements. Yet, to operationalize, this annotation-based approach can be costly and time-consuming, especially when dealing with a large number of diverse queries. On the other hand, \citet{Knowledge-to-SQL} propose an automatic generation of knowledge, based on the question and its associated database schema. However, this method is still limiting as it generates only a few pieces of knowledge for each query without leveraging the potential for reuse. In contrast, since much of the knowledge used for one query can be applicable to multiple similar queries (See Figure~\ref{fig:concept}, Right), we aim to design a more effective approach for knowledge augmentation, discussed below.

\subsection{Knowledge Base Construction}

To address the aforementioned limitations of existing approaches in knowledge augmentation for text-to-SQL, we propose a novel approach to automatically construct a comprehensive and reusable knowledge base. Ideally, this can serve as a foundational resource, encapsulating diverse domain information and offering insights into various database schemas, to enhance the understanding of queries and their associated database structures.

Formally, we design this knowledge base $\mathcal{K}$ as a collection of knowledge entries, each represented as a concise sentence, denoted as follows: $\vk \in \mathcal{K}$. For instance, in the medical domain, one knowledge entry might be ``Abnormal white blood cell count refers to WBC $\leq$ 3.5 or WBC $\geq$ 9.0'', which describes the abnormal range of white blood cell counts and its corresponding column name ``WBC'' in the database schema, applicable to queries related to abnormal white blood cells. The next question to answer is then how to construct this knowledge base based on the available resources. 

In this work, we start with collecting all the existing knowledge entries from the publicly available dataset~\cite{bird}, which includes the knowledge and its related pair of query and database schema. Yet, while this initial collection can serve as the foundational layer of our knowledge base, it may not capture the full scope of the required information. To address this gap, we propose an automatic knowledge base expansion technique that leverages LLMs, which possess domain-specific knowledge and the ability to comprehend the given context (including instructions, codes, and database structures) by generating additional knowledge entries. Specifically, given the query and its associated database schema from the available datasets, we prompt LLMs (along with a prompting template $\mathcal{T}$ for knowledge generation) to produce the knowledge, formulated as follows: $\vk = \texttt{LLM}(\mathcal{T}(\vq, \mathcal{D}))$, and then store this knowledge $\vk$ into the knowledge base $\mathcal{K}$. In addition, as it may be more accurate and reliable to provide the LLM with relevant examples (which can help it understand the context, nuances, and expectations of the desired output), we further prepend the small number of relevant examples into the prompt of $\texttt{LLM}$. It is worth noting that these examples are comprised of the triplets of the user queries, their associated database schemas, and the knowledge they are derived from, and that those triplets come from the existing dataset (used to construct the initial knowledge base). Also, we select only those highly relevant to the query based on its embedding-level cosine similarities with samples from the existing dataset, calculated by MPNet~\cite{MPNet}. This process can ultimately enable the LLM to generate more precise and contextually appropriate knowledge for text-to-SQL.

In addition to this relevant example-based knowledge generation approach, to further enrich the diversity and comprehensiveness of the knowledge base, we implement a simple yet effective strategy that involves sampling and permutation of few-shot examples provided to the LLM. Specifically, for the given query and its associated database schema, instead of generating their corresponding knowledge only once, we iteratively sample a different set of relevant examples (provided to contextualize the LLM) multiple times and further permute their order. This can allow the LLM to explore different contextual nuances and generate a wider range of knowledge entries, with the goal of ultimately increasing the robustness and applicability of the knowledge base for a broader range of queries.

\input{Algorithms/overall}

\subsection{Text-to-SQL with Knowledge Base}

Based on the LLM-powered knowledge base construction process, we now have the knowledge base $\mathcal{K}$. Hereafter, the next question to answer is then how to use this knowledge base for text-to-SQL. 

Given the extensive nature of $\mathcal{K}$, containing a large number of entries, it is crucial to identify and retrieve the most pertinent entries for the query $\vq$. Formally, this process can be represented as follows: $\{\vk_i\}_{i=1}^j = \texttt{Retriever}(\vq, \mathcal{K})$. Also, this can be operationalized by calculating the embedding-level similarities between the query and all the knowledge entries in the knowledge base, then selecting the top-$j$ similar entries $\{\vk_i\}_{i=1}^j$, where embeddings are obtained from a sentence embedding model~\cite{dpr, MPNet}. Moreover, to further enhance the retrieval accuracy, we train this embedding model with contrastive learning, which maximizes the similarity between the query and its relevant knowledge while minimizing the similarities of others, denoted as follows: $-\log \frac{\exp(\text{sim}(\vq, \vk^+) / \tau)}{\exp(\text{sim}(\vq, \vk^+) / \tau) + \sum_{\vk^-} \exp(\text{sim}(\vq, \vk^-) / \tau)}$, where $\text{sim}(\vq, \vk)$ denotes the similarity measure between query $\vq$ and knowledge $\vk$, $\tau$ is the temperature parameter, $\vk^+$ is the relevant knowledge, and $\vk^-$ represents the set of irrelevant knowledge.

Note that while the retrieved knowledge entries from $\mathcal{K}$ are relevant to the given query and can assist in SQL statement formulation, they may require additional refinement to perfectly align with the query's specific needs. For instance, if the user query pertains to abnormal data conditions, but the retrieved knowledge primarily focuses on normal data, a direct application of this knowledge could lead to inaccurate SQL generation. To address this issue, we further prompt the LLM to generate the knowledge tailored to the given query by considering its relevant knowledge entries and database schema, as follows: $\vk' = \texttt{LLM}(\mathcal{T}(\vq, \{\vk_i\}_{i=1}^j, \mathcal{D}))$, where $\{\vk_i\}_{i=1}^j$ is the knowledge retrieved from $\mathcal{K}$. This refined knowledge $\vk'$ is subsequently used as input, along with the user query and its associated database schema, to guide the text-to-SQL LLM in generating a more accurate and contextually appropriate SQL statement: $\vs = \texttt{LLM}(\mathcal{T}(\vq, \vk', \mathcal{D}))$. Please see Algorithm~\ref{alg:kat-sql-overall} for our overall approach.

%% file: Algorithms/overall.tex
\begin{figure}[t]
\vspace{-0.1in}
\begin{algorithm}[H]
\small
\caption{Knowledge-Augmented Text-to-SQL}
\label{alg:kat-sql-overall}
\begin{spacing}{1.0}
\begin{algorithmic}[1]
\Require Dataset $\mD$ containing query-schema pairs $(\vq, \mathcal{D})$; LLM model $\texttt{LLM}$; Prompt templates $\mathcal{T}$
\Ensure SQL statement $\vs$ for a given query $\vq$

\vspace{0.025in}
\State \textbf{Phase 1: Knowledge Base Construction}
\State $\mathcal{K} \gets \{\} \cup \mD$ \Comment{Initialize knowledge base}
\ForAll{$(\vq, \mathcal{D}) \in \mD$}
    \State $\mathcal{E} \gets$ Retrieve top-$k$ relevant examples from $\mD$
    \State $\vk_{\text{new}} \gets \texttt{LLM}(\mathcal{T}_{\text{gen}}(\vq, \mathcal{D}, \mathcal{E}))$ \Comment{Generate knowledge}
    \State $\mathcal{K} \gets \mathcal{K} \cup \vk_{\text{new}}$ \Comment{Store knowledge}
\EndFor

\vspace{0.025in}
\State \textbf{Phase 2: Knowledge-Augmented SQL Generation}
\Function{KAT-SQL}{$\vq$, $\mathcal{D}$, $\mathcal{K}$}
    \State $\{\vk_i\}_{i=1}^j \gets$ Retrieve top-$j$ knowledge from $\mathcal{K}$
    \State $\vk' \gets \texttt{LLM}(\mathcal{T}_{\text{ref}}(\vq, \{\vk_i\}_{i=1}^j, \mathcal{D}))$ \Comment{Refine knowledge}
    \State $\vs \gets \texttt{LLM}(\mathcal{T}_{\text{text-to-SQL}}(\vq, \vk', \mathcal{D}))$ \Comment{Generate SQL}
    \State \Return $\vs$
\EndFunction
\end{algorithmic}
\end{spacing}
\end{algorithm}
\vspace{-0.225in}
\caption{A simplified overview of the proposed KAT-SQL method. Please see Algorithms~\ref{alg:knowledge_base_construction} and \ref{alg:kat-sql} for detailed versions.}
\vspace{-0.1in}
\end{figure}

%% file: Sections/5_experimental_setup.tex
\section{Experimental Setup}

\subsection{Datasets and Tasks}

\input{Tables/main_results}

\paragraph{Datasets}
To validate the efficacy of KAT-SQL, we first use two widely used text-to-SQL benchmark datasets, namely BIRD~\cite{bird} and Spider~\cite{Spider}. Specifically, BIRD is a recently released large-scale text-to-SQL dataset, built on top of 95 distinct databases spanning 37 domains. Additionally, each query in this dataset is associated with knowledge that is manually annotated by humans, providing a useful prior for formulating SQL statements. Spider is another benchmark dataset, built upon 200 databases across 138 domains. Unlike BIRD, samples in Spider do not have annotated knowledge for text-to-SQL. Lastly, we consider a challenging real-world text-to-SQL data, namely CSTINSIGHT, which is designed with actual customer queries over a data lakehouse with 34 tables without human-annotated knowledge.

\paragraph{Tasks/Scenarios}
We evaluate our KAT-SQL on three realistic text-to-SQL tasks. First of all, we consider the scenario where the prior information about some samples and their associated knowledge for each database is available, meaning that the databases used in training samples overlap with those in test samples (Overlap). We note that this setting is practical, since annotating a few pairs of questions and their corresponding knowledge for each database in advance is feasible. In addition to this, we test KAT-SQL with the existing benchmark setup, which is more challenging since it assumes there are no overlaps between databases during the training and test phases (Non-Overlap). In other words, no samples from the test-time databases are available beforehand, which means the model should be able to generalize to test-time queries based on the schemas of test-time databases as well as the samples and their associated knowledge from the different (training-time) databases. Lastly, we validate KAT-SQL on the most challenging scenario, where there are no overlaps between the databases used during training and testing, but also no knowledge is available for both training and test samples. This setup aims to test the model's ability to generalize (in the absence of any prior knowledge about the dataset), allowing us to evaluate how well our knowledge base constructed with one dataset performs on different datasets. Notably, since the Spider and CSTINSIGHT datasets have no available knowledge for all queries, we use them for the most challenging last scenario; meanwhile, we use the BIRD dataset for the first two scenarios.

\subsection{Baselines and Our Model}

We compare our KAT-SQL approach against relevant baselines that target our primary objective of improving knowledge-augmented text-to-SQL systems, which vary in their usage of knowledge. We note that for the fairest comparison, we fix the LLM as the same for all methods, explained as follows:
\begin{enumerate*}[itemsep=0.0mm, parsep=1pt, leftmargin=*]
\item{\bf No Knowledge} -- which uses only the queries themselves to formulate the SQL statements without any additional knowledge.
\item{\bf DELLM} -- which generates the knowledge based on the query and its relevant database structures, and use this synthesized knowledge for text-to-SQL~\cite{Knowledge-to-SQL}.
\item{\bf KAT-SQL} -- which is our model, building the knowledge base and utilizing the knowledge from it (with retrieval) for text-to-SQL.
\item{\bf Oracle Knowledge} -- which uses oracle knowledge annotated by humans, along with the queries to generate the SQL statements. This approach serves as an upper bound and is not directly comparable to other models due to its reliance on accurate, manually curated knowledge that is typically unavailable.
\end{enumerate*}

\subsection{Evaluation Metrics}

Following the standard evaluation protocols from prior work~\cite{bird, Knowledge-to-SQL}, we use the following two metrics: 1) Execution Accuracy (EX), which measures the ratio of generated SQL code that has the same execution results with ground-truth SQL code; 2) Valid Efficiency Score (VES), which considers the efficiency of generated SQLs by weighting them based on their relative efficiency improvement over ground-truth SQLs further multiplied by execution accuracy. 

\subsection{Implementation Details}

We mainly use Llama-3 70B~\cite{llama3modelcard} as the basis for text-to-SQL generation and knowledge generation across all baselines and our model variants for most experiments, for a fair comparison, while we also experiment with other LLMs in an analysis (Table~\ref{tab:different_llms}) to see the robustness of KAT-SQL. For the hyperparameters, except for the temperature (which we set as 0.0 for reproducibility), we use its default values. In addition, for the retriever, we use MPNet~\cite{MPNet}, which is based on dense retrieval; we train it with a batch size of 128 and a number of training epochs of 30. We provide the detailed prompts used to elicit the knowledge and SQL generations in Appendix~\ref{sec:appendix:prompt}.

%% file: Tables/main_results.tex
\begin{table*}[t!]
\caption{Main results on text-to-SQL benchmark datasets across multiple scenarios, with the best results in bold.}
\vspace{-0.1in}
\label{tab:main}
\small
\centering
\resizebox{\textwidth}{!}{
\renewcommand{\arraystretch}{1.2}
\renewcommand{\tabcolsep}{5.0mm}
\begin{tabular}{lcccccccc}
\toprule

& \multicolumn{2}{c}{\bf BIRD (Overlap)} & \multicolumn{2}{c}{\bf BIRD (Non-Overlap)} & \multicolumn{2}{c}{\bf Spider} & \multicolumn{2}{c}{\bf CSTINSIGHT} \\
\cmidrule(l{2pt}r{2pt}){2-3} \cmidrule(l{2pt}r{2pt}){4-5} \cmidrule(l{2pt}r{2pt}){6-7} \cmidrule(l{2pt}r{2pt}){8-9} 
\textbf{Methods} & EX & VES & EX & VES & EX & VES & EX & VES \\

\midrule
\midrule

\textbf{No Knowledge} & 23.76 & 28.81 & 20.66 & 16.72 & 70.99 & 37.53 & 4.76 & 5.28 \\

\textbf{DELLM} & 34.70 & 33.15 & 24.64 & 19.27 & 72.44 & 42.90 & 11.90 & 12.02 \\

\textbf{KAT-SQL (Ours)} & \textbf{41.18} & \textbf{41.33} & \textbf{41.07} & \textbf{31.14} & \textbf{74.56} & \textbf{47.20} & \textbf{14.29} & \textbf{14.50} \\

\noalign{\vskip 0.25ex}\cdashline{1-9}\noalign{\vskip 0.75ex}

\textbf{Oracle Knowledge} & 54.67 & 49.71 & 49.41 & 37.93 & N/A & N/A & N/A & N/A \\

\bottomrule

\end{tabular}
}
\vspace{-0.1in}
\end{table*}

%% file: Sections/6_experimental_result.tex
\section{Experimental Results and Analyses}

\input{Figures/knowledge_construction}

\paragraph{Main Results}
We provide main results in Table~\ref{tab:main}, which confirms that our KAT-SQL approach consistently outperforms all baselines by large margins. Specifically, while we observe some performance improvement of the knowledge-augmented text-to-SQL approach (namely DELLM, which generates a few pieces of knowledge for each query) over the baseline without knowledge augmentation, KAT-SQL achieves even greater gains, demonstrating the effectiveness of our knowledge base construction-based text-to-SQL paradigm. However, the performance of the (incomparable) model with the oracle knowledge (annotated by human experts) remains superior to all other approaches, which suggests potential future opportunities for developing a more advanced pipeline for knowledge generation.

\paragraph{Analysis on Knowledge Base}
To further understand the coverage and relevance of the knowledge within our knowledge base, we compare each piece of knowledge required for test-time queries with all the available entries in the knowledge base, as a function of the number of knowledge generation steps during knowledge base construction. For evaluation, we use two metrics: Exact Match, which identifies whether the knowledge base contains an entry that precisely matches the knowledge required for a given query, and Semantic Similarity, which assesses how closely related the most similar entry (in the knowledge base) is to the required knowledge based on the embedding-level similarity. As shown in Figure~\ref{fig:construction}, we observe that, under the Overlap setting, half of the knowledge entries needed for test-time queries are available in the knowledge base while the Semantic Similarity is around 90\%, which demonstrates substantial coverage by our knowledge base. In addition, for the challenging setup where training and test databases are distinct, we still observe that 20\% of the test-time knowledge entries are available in the knowledge base and that the Semantic Similarity exceeds 80\%, showing the utility of our knowledge base. Finally, as we increase the number of knowledge generation steps for each instance during knowledge base construction, we observe a corresponding improvement in both coverage and relevance of our knowledge base, which supports the effectiveness of 
our expansion strategy to enrich its diversity.

\input{Tables/knowledge_generation}

\paragraph{Analysis on Knowledge Generation}
Recall that we further refine the retrieved knowledge to make it more suitable for each query, in addition to constructing the knowledge base and retrieving the relevant knowledge. Thus, to see how relevant the generated knowledge is to the human-annotated gold knowledge with regards to the use of our knowledge base, we report comparison results according to Exact Match and Semantic Similarity (SS) in Table~\ref{tab:knowledge_generation}. We observe that when we retrieve the relevant knowledge from the knowledge base and then use it for knowledge generation, there are performance gains over the case where we do not leverage it, which indicates that the retrieved knowledge is helpful in formulating the necessary knowledge for test-time queries. We also provide few-shot examples to guide the knowledge generation model in generating useful knowledge in the right format, and when we select them based on their similarities with the given query, we observe further gains in the quality of the generated knowledge.

Beyond evaluating the quality of the generated knowledge by comparing it to the human-annotated gold knowledge, we also examine the impact of knowledge generation on downstream text-to-SQL performance with and without the incorporation of generated knowledge. As shown in Table~\ref{tab:knowledge_generation_ablation}, compared to the results without the knowledge retrieval and generation on both Overlap and Non-Overlap settings, there are substantial improvements when we incorporate the retrieved knowledge from our knowledge base into the text-to-SQL generation process. Furthermore, instead of directly using the retrieved knowledge, refining this retrieved knowledge yields additional improvements, underscoring the importance of not only retrieving relevant knowledge but also tailoring it to better align with the specific needs of test-time queries.

\input{Tables/knowledge_generation_ablation}
\input{Tables/knowledge_retrieval}

\paragraph{Retrieval Analysis}
We also analyze the accuracy of knowledge retrieval from our knowledge base by reporting its retrieval performance in Table~\ref{tab:knowledge_retrieval} according to Mean Reciprocal Rank (MRR) and Top@K Accuracy. We observe that the retrieval accuracy on the Overlap setting is higher than that on the Non-Overlap setting, due to the less availability of relevant knowledge required for test-time queries in the Non-Overlap setting. Yet, when we replace the knowledge base constructed from our approach with the Oracle knowledge base (*), which includes all the necessary knowledge for test-time queries, the MRR on both settings reaches around 80\%, indicating the importance of expanding the coverage of the knowledge base for accurate knowledge retrieval. The table also compares the performance of different basis models for retrieval -- BERT~\cite{BERT} and TAS-B~\cite{TAS-B} -- with the latter being fine-tuned for retrieval. It can be seen that the extra training of the model on retrieval tasks aids in achieving superior performance for retrieving the knowledge for text-to-SQL.

\input{Tables/domain_generalization}

\paragraph{Generalization Analysis to Different Domains}
To see whether our knowledge base can be generalizable to databases of different domains (that are not overlapped with those for knowledge base construction), we breakdown the performance based on whether test databases share domains with training databases or belong to different domains (according to 37 domains categorized from~\citet{bird}). As shown in Table~\ref{tab:domain_generalization}, our KAT-SQL achieves substantially higher performance when test databases overlap with training domains compared to those from unseen domains; however, even in the latter case, KAT-SQL still outperforms existing baselines. These results indicate that, while the lack of domain overlaps degrades the performance, our knowledge base still provides meaningful benefits for unseen domains, demonstrating its generalizability.

\paragraph{Analysis with Different LLMs}
To evaluate how robust our KAT-SAL approach is across different LLMs, we conduct the additional analysis instantiating the text-to-SQL and knowledge generation models with other recent LLMs such as Granite 34B~\cite{Granite} and Mixtral 8x7B~\cite{Mixtral};  results are shown in Table~\ref{tab:different_llms}. From this, we observe that KAT-SQL consistently outperforms all baselines regardless of the choice of LLMs, which demonstrates the effectiveness and versatility of our proposed approach. 

Finally, we augment the state-of-the-art text-to-SQL model (in the setting without oracle knowledge) on the BIRD leaderboard~\cite{bird}, namely ExSL + granite-20b-code, using the knowledge generated from our proposed knowledge base construction-based approach. As shown in Table~\ref{tab:leaderboard}, we observe that the text-to-SQL model combined with our KAT-SQL approach establishes the new state-of-the-art performance, highlighting the value of our method as a powerful tool for text-to-SQL.

\input{Tables/different_llms}
\input{Tables/leaderboard}

\input{Tables/case_study}
\input{Tables/example_knowledge}

\paragraph{Analysis on Efficiency}
While our primary focus is on improving the text-to-SQL accuracy through knowledge base construction and augmentation, we also consider the efficiency of our approach. It is worth noting that the construction of the knowledge base is performed offline and does not affect real-time query processing; therefore, the extra computational overhead comes from retrieving relevant knowledge and generating the SQL statement in response to the query. In this regard, our retrieval process accounts for only 2\% of the overall generation time, thanks to efficient search algorithm~\cite{faiss}, making its impact negligible. Also, although incorporating knowledge into the text-to-SQL pipeline increases the prompt length by 30\%, this overhead aligns with other knowledge augmentation methods (such as DELLM) and does not introduce additional latency specific to our approach. Overall, each query is processed under 5 seconds.

\paragraph{Examples} 
We provide examples for the knowledge generation and text-to-SQL in Table~\ref{tab:case_study} as well as the entries in the knowledge base in Table~\ref{tab:example_knowledge}. 

%% file: Figures/knowledge_construction.tex
\begin{figure}[t!]
    \centering
    \vspace{-0.025in}
    \includegraphics[width=0.975\columnwidth]{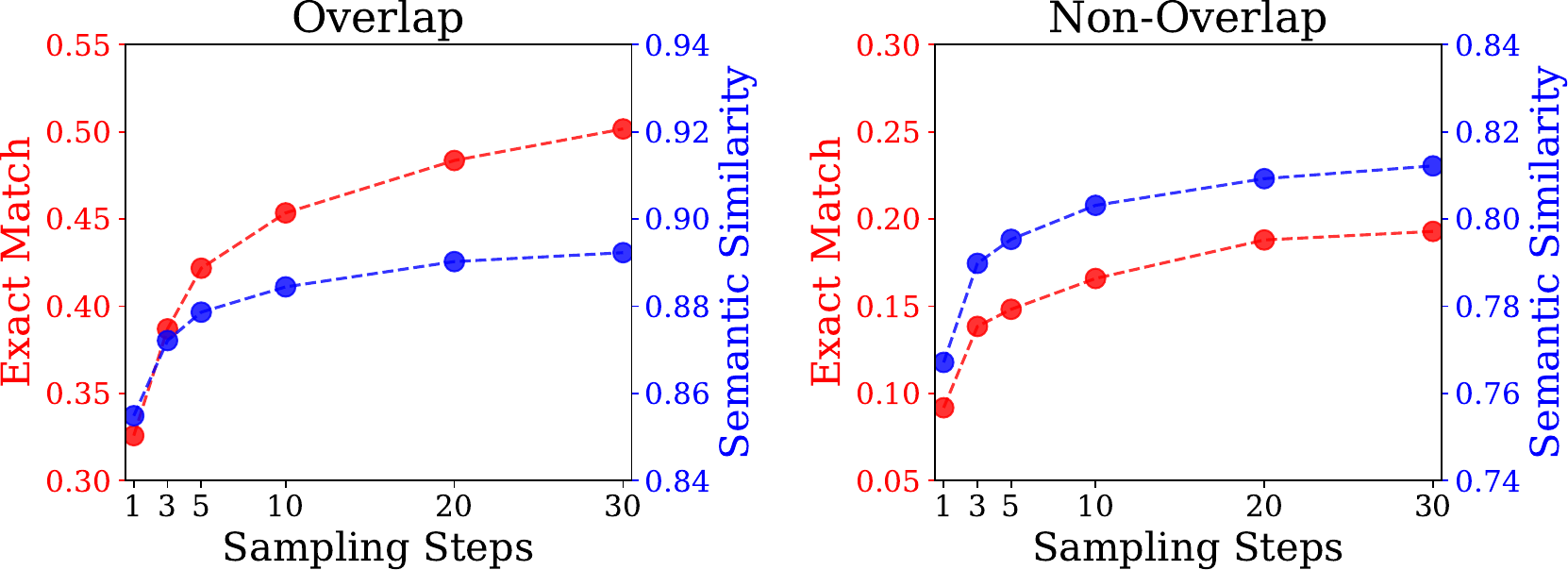}
    \vspace{-0.1in}
    \caption{Results for coverage and relevance of knowledge entries in the constructed knowledge base against gold knowledge, with different numbers of knowledge generation steps.}
    \label{fig:construction}
    \vspace{-0.1in}
\end{figure}

%% file: Tables/knowledge_generation.tex
\begin{table}[t!]
\caption{Results for knowledge generation with and without the use of the Knowledge Base (KB), while varying the prompt construction with and without the relevant few-shot examples.}
\vspace{-0.1in}
\label{tab:knowledge_generation}
\small
\centering
\resizebox{0.475\textwidth}{!}{
\renewcommand{\arraystretch}{1.0}
\renewcommand{\tabcolsep}{3.0mm}
\begin{tabular}{llcccc}
\toprule

& & \multicolumn{2}{c}{\bf Overlap} & \multicolumn{2}{c}{\bf Non-Overlap} \\
\cmidrule(l{2pt}r{2pt}){3-4} \cmidrule(l{2pt}r{2pt}){5-6} 
\textbf{KB} & \textbf{Few-Shot} & \textbf{EM} & \textbf{SS} & \textbf{EM} & \textbf{SS} \\

\midrule
\midrule

\multirowcell{2}[-0.0ex][l]{\textbf{w/o KB}} 

& Random & 10.96 & 68.77 & 7.88 & 66.77  \\

& Retrieval & 20.21 & 73.62 & 9.24 & 68.78 \\

\midrule

\multirowcell{2}[-0.0ex][l]{\textbf{w/ KB}} 

& Random & 11.13 & 69.14 & 7.93 & 66.80 \\

& Retrieval & \textbf{24.97} & \textbf{77.87} & \textbf{12.94} & \textbf{71.24} \\

\bottomrule

\end{tabular}
}
\vspace{-0.1in}
\end{table}

%% file: Tables/knowledge_generation_ablation.tex
\begin{table}[t!]
\caption{Text-to-SQL results without using any knowledge, based on the retrieved knowledge, and based on the refined knowledge from the retrieved knowledge (Our KAT-SQL).}
\vspace{-0.1in}
\label{tab:knowledge_generation_ablation}
\small
\centering
\resizebox{0.475\textwidth}{!}{
\renewcommand{\arraystretch}{1.0}
\renewcommand{\tabcolsep}{3.0mm}
\begin{tabular}{llc}
\toprule

\textbf{Settings} & \textbf{Models} & \textbf{EX} \\

\midrule
\midrule

\multirowcell{3}[-0.5ex][l]{\textbf{Overlap}} 

& KAT-SQL (Ours) & 41.18 \\

\noalign{\vskip 0.25ex}\cdashline{2-3}\noalign{\vskip 0.75ex}

& w/o Generation & 38.94 \\

& w/o Retrieval \& Generation & 23.76 \\

\midrule

\multirowcell{3}[-0.5ex][l]{\textbf{Non-Overlap}} 

& KAT-SQL (Ours) & 41.07 \\

\noalign{\vskip 0.25ex}\cdashline{2-3}\noalign{\vskip 0.75ex}

& w/o Generation & 38.42 \\

& w/o Retrieval \& Generation & 20.66 \\

\bottomrule

\end{tabular}
}
\end{table}

%% file: Tables/knowledge_retrieval.tex
\begin{table}[t!]
\caption{Retrieval results with different scenarios and models.}
\vspace{-0.1in}
\label{tab:knowledge_retrieval}
\small
\centering
\resizebox{0.475\textwidth}{!}{
\renewcommand{\arraystretch}{0.8}
\renewcommand{\tabcolsep}{3.0mm}
\begin{tabular}{llccc}
\toprule

\textbf{Settings} & \textbf{Models} & \textbf{MRR} & \textbf{Top@3} & \textbf{Top@10} \\

\midrule
\midrule

\multirowcell{3}[-0.5ex][l]{\textbf{Overlap}} 

& BERT & 0.5506 & 0.6621 & 0.8911 \\

& TAS-B & 0.5630 & 0.6943 & 0.9035 \\

\noalign{\vskip 0.25ex}\cdashline{2-5}\noalign{\vskip 0.75ex}

& TAS-B* & 0.8288 & 0.9143 & 0.9765 \\

\midrule

\multirowcell{3}[-0.5ex][l]{\textbf{Non-Overlap}} 

& BERT & 0.2148 & 0.2692 & 0.4231 \\

& TAS-B & 0.2364 & 0.3846 & 0.4615 \\

\noalign{\vskip 0.25ex}\cdashline{2-5}\noalign{\vskip 0.75ex}

& TAS-B* & 0.7565 & 0.8347 & 0.9210 \\

\bottomrule

\end{tabular}
}
\vspace{-0.1in}
\end{table}

%% file: Tables/domain_generalization.tex
\begin{table}[t!]
\caption{Breakdown text-to-SQL results into overlapping and non-overlapping domain settings between training (knowledge base construction) and test (text-to-SQL evaluation) databases. }
\vspace{-0.1in}
\label{tab:domain_generalization}
\small
\centering
\resizebox{0.475\textwidth}{!}{
\renewcommand{\arraystretch}{1.0}
\renewcommand{\tabcolsep}{5.0mm}
\begin{tabular}{lcc}
\toprule

\textbf{Models} & \textbf{Overlap} & \textbf{Non-Overlap} \\

\midrule
\midrule

No Knowledge & 22.85 & 16.20 \\

DELLM & 27.20 & 19.43 \\

\noalign{\vskip 0.25ex}\cdashline{1-3}\noalign{\vskip 0.75ex}

KAT-SQL (Ours) & \textbf{49.37} & \textbf{24.19} \\

\bottomrule

\end{tabular}
}
\vspace{-0.1in}
\end{table}

%% file: Tables/different_llms.tex
\begin{table}[t!]
\caption{Text to SQL results with different LLMs.}
\vspace{-0.1in}
\label{tab:different_llms}
\small
\centering
\resizebox{0.475\textwidth}{!}{
\renewcommand{\arraystretch}{0.75}
\renewcommand{\tabcolsep}{3.0mm}
\begin{tabular}{llcc}
\toprule

\textbf{LLMs} & \textbf{Methods} & \textbf{Overlap} & \textbf{Non-Overlap} \\

\midrule
\midrule

\multirowcell{4}[-0.5ex][l]{\textbf{Llama}} 

& No Knowledge & 23.76 & 20.66 \\

& DELLM & 34.70 & 24.64 \\

& KAT-SQL & 41.18 & 41.07 \\

\noalign{\vskip 0.25ex}\cdashline{2-4}\noalign{\vskip 0.75ex}

& Oracle Knowledge & 54.67 & 49.41 \\

\midrule

\multirowcell{4}[-0.5ex][l]{\textbf{Granite}} 

& No Knowledge & 25.83 & 17.75 \\

& DELLM & 34.04 & 20.21 \\

& KAT-SQL & 39.28 & 35.83 \\

\noalign{\vskip 0.25ex}\cdashline{2-4}\noalign{\vskip 0.75ex}

& Oracle Knowledge & 46.56 & 38.32 \\

\midrule

\multirowcell{4}[-0.5ex][l]{\textbf{Mixtral}} 

& No Knowledge & 11.75 & 10.58 \\

& DELLM & 27.17 & 11.29 \\

& KAT-SQL & 29.31 & 20.30 \\

\noalign{\vskip 0.25ex}\cdashline{2-4}\noalign{\vskip 0.75ex}

& Oracle Knowledge & 37.26 & 30.88 \\

\bottomrule

\end{tabular}
}
\end{table}

%% file: Tables/leaderboard.tex
\begin{table}[t!]
\caption{Results of our KAT-SQL approach with the state-of-the-art text-to-SQL model on the BIRD leaderboard.}
\vspace{-0.1in}
\label{tab:leaderboard}
\small
\centering
\resizebox{0.475\textwidth}{!}{
\renewcommand{\arraystretch}{1.0}
\renewcommand{\tabcolsep}{3.0mm}
\begin{tabular}{lc}
\toprule

\textbf{Models} & \textbf{EX} \\

\midrule
\midrule

ChatGPT & 24.05 \\

ChatGPT + CoT & 25.88 \\

ExSL + granite-20b-code & 51.69 \\

ExSL + granite-20b-code w/ KAT-SQL (Ours) & 57.56 \\

\noalign{\vskip 0.25ex}\cdashline{1-2}\noalign{\vskip 0.75ex}

ExSL + granite-20b-code w/ Oracle Knowledge & 65.38 \\

\bottomrule

\end{tabular}
}
\vspace{-0.1in}
\end{table}

%% file: Tables/case_study.tex
\begin{table*}[tbp]
\centering
\caption{Case study of knowledge and SQL generation results from three queries.}
\vspace{-0.1in}
\label{tab:case_study}
\small
\adjustbox{max width=0.875\textwidth}{
\begin{tabularx}{\textwidth}{|m{1.4cm}|m{3.6cm}|m{\dimexpr\textwidth-5cm-6\tabcolsep\relax}|}
\hline
\multicolumn{1}{|c|}{\textbf{Example}} & \multicolumn{1}{c|}{\textbf{Query}} & \multicolumn{1}{c|}{\textbf{Generated Knowledge and SQL}} \\ \hline

\textbf{Example 1} & 
Which position has a lower minimum salary, Account Representative or Trainee? & 
\textbf{Generated Knowledge:} \newline
1) lower minimum salary refers to \texttt{MIN(minsalary)} \newline
2) Account Representative and Trainee are position titles \newline
\textbf{Generated SQL:} \newline
\texttt{SELECT positiontitle FROM position WHERE positiontitle = 'Account Representative' OR positiontitle = 'Trainee' ORDER BY minsalary ASC LIMIT 1} \\ \hline

\textbf{Example 2} & 
Among the employees working at the office in New York, how many of them have a good job performance? & 
\textbf{Generated Knowledge:} \newline
1) New York refers to \texttt{state = 'NY'} \newline
2) good job performance refers to \texttt{performance = 'Good'} \newline
\textbf{Generated SQL:} \newline
\texttt{SELECT COUNT(*) FROM employee AS T1 INNER JOIN location AS T2 ON T1.locationID = T2.locationID WHERE T2.state = 'NY' AND T1.performance = 'Good'} \\ \hline

\textbf{Example 3} & 
For all patients born in 1982, state if their albumin is within normal range. & 
\textbf{Generated Knowledge:} \newline
1) albumin is within normal range refers to \texttt{ALB between 3.5 and 5.5} \newline
\textbf{Generated SQL:} \newline
\texttt{SELECT CASE WHEN T2.ALB >= 3.5 AND T2.ALB <= 5.5 THEN 'normal' ELSE 'abnormal' END FROM Patient AS T1 INNER JOIN Laboratory AS T2 ON T1.ID = T2.ID WHERE STRFTIME('\%Y', T1.Birthday) = '1982'} \\ \hline

\end{tabularx}
}
\end{table*}

%% file: Tables/example_knowledge.tex
\begin{table*}[tbp]
\centering
\caption{Examples of original and its (similar) constructed knowledge within the knowledge base.}
\label{tab:example_knowledge}
\vspace{-0.1in}
\small
\adjustbox{max width=0.875\textwidth}{
\begin{tabularx}{\textwidth}{|m{1.4cm}|m{3.6cm}|m{\dimexpr\textwidth-5cm-6\tabcolsep\relax}|}
\hline
\multicolumn{1}{|c|}{\textbf{Example}} & \multicolumn{1}{c|}{\textbf{Original Knowledge}} & \multicolumn{1}{c|}{\textbf{Constructed Similar Knowledge}} \\ \hline

\textbf{Example 1} & 
albumin is within normal range refers to \texttt{ALB between 3.5 and 5.5} & 
1) albumin is outside the normal range refers to \texttt{ALB less than 3.5 or greater than 5.5} \newline
2) glucose is within normal range refers to \texttt{GLU between 70 and 100 mg/dL} \newline
3) Hemoglobin (Hb) is considered normal for males if levels range from \texttt{13.5 to 17.5 g/dL} \\ \hline

\textbf{Example 2} & 
Eligible free rate for K-12 = Free Meal Count (K-12) / Enrollment (K-12) & 
1) Eligible reduced-price rate for K-12 = Reduced-Price Meal Count (K-12) / Enrollment (K-12) \newline
2) Eligible free meal rate for students aged 5-17 = Free Meal Count (Ages 5-17) / Enrollment (Ages 5-17) \newline
3) Difference between K-12 and ages 5-17 enrollment = \texttt{Enrollment (K-12) - Enrollment (Ages 5-17)} \\ \hline

\textbf{Example 3} & 
Slovakia can be represented as \texttt{Country = 'SVK'} & 
1) France can be represented as \texttt{Country = 'FRA'} \newline
2) Brazil can be represented as \texttt{Country = 'BRA'} \newline
3) Monaco can be represented as \texttt{Country = 'MCO'} \\ \hline

\end{tabularx}
}
\end{table*}

%% file: Sections/7_conclusion.tex
\section{Conclusion}
In this work, we proposed a novel knowledge base construction-based text-to-SQL approach called KAT-SQL, based on the motivation that one piece of knowledge can be reused across multiple queries and databases. Our approach involves the creation of the knowledge base from which relevant knowledge is retrieved and utilized to generate SQL statements from queries. Through extensive evaluations on multiple datasets with two different scenarios, we showed that our KAT-SQL outperforms relevant knowledge-augmented text-to-SQL baselines. In addition, our detailed analyses highlight the effectiveness of each component in the knowledge generation and retrieval processes, but also the high coverage and relevance of the entries in the base. 

\section*{Limitations}
In this work, we propose constructing a knowledge base and then leveraging it for text-to-SQL tasks, showcasing the clear advantages of constructing the knowledge base for text-to-SQL. However, as the performance gaps between the models with oracle knowledge and the generated knowledge from our knowledge base indicate, there is still room to improve the coverage of the knowledge base with advanced knowledge base construction methods, which is a promising area for future work.

\section*{Ethics Statement}
We recognize that any text-to-SQL system, including our proposed approach, may carry the inherent risk of generating SQL queries that may inadvertently or intentionally access, modify, or delete sensitive information within a database. While this vulnerability is not exclusive to our method and is a well-known challenge in the broader field of text-to-SQL systems, it underscores the importance of implementing robust security measures and access controls before deploying such systems. Similar to this, safety is particularly crucial in our application, so as to avoid the risk of sensitive information being stored in the knowledge base and subsequently being inappropriately reused.

%% file: Sections/8_appendix.tex
\clearpage

\appendix

\section{Prompts}
\label{sec:appendix:prompt}

We provide prompts used to elicit the knowledge generation and the SQL generation in Table~\ref{tab:prompt}.

\section{Algorithms}

We provide the pseudo-code for knowledge base construction in Algorithm~\ref{alg:knowledge_base_construction} and the pseudo-code for our full KAT-SQL approach in Algorithm~\ref{alg:kat-sql}.

\section{Additional Experimental Results}

\paragraph{Knowledge Base Statistics}
The resulting knowledge base for the database overlapping and non-overlapping scenarios contains 86,254 and 117,328 knowledge entries, respectively, which are greater than the original number of knowledge entries annotated in the BIRD dataset, which is 12,751.

\paragraph{Knowledge Base Construction Cost}
While the construction of the knowledge base is performed offline and does not impact real-time operations of text-to-SQL, we provide the cost to construct the knowledge base for our KAT2SQL approach to enable researchers to estimate resource requirements for scaling and implementation. Note that the exact computational costs and time required for knowledge base construction vary depending on hardware types and configurations, and with four H100 GPUs that can process 2K tokens per second and generate 10 tokens per second for Llama 70B, the time required to generate each knowledge entry is around 2 seconds. Therefore, for the knowledge base with 100K entries, the total generation time would be 56 hours divided by the number of parallel models (it completes in 7 hours with 8 models).

\paragraph{Retrieval over Different Knowledge Sources}
It is worthwhile to note that, for text-to-SQL tasks, it is crucial to consider the relationship between the query and the database (in addition to the consideration of the domain-specific knowledge for domain-specific queries); therefore, using the unstructured knowledge sources (such as web search) may not be optimal for this purpose since they often lack the structured, schema-specific information necessary for accurately formulating SQL queries. Nevertheless, to further validate this claim, we perform retrieval over Wikipedia, instead of performing retrieval over the constructed knowledge base, and observe only the marginal performance gain (3\%) compared to the baseline without augmentation.

\input{Tables/prompt}

\clearpage

\input{Algorithms/KB-Construction}
\input{Algorithms/KAT-SQL}

%% file: Tables/prompt.tex
\begin{table*}[t!]
    \caption{A list of prompts that we use for knowledge generation and SQL generation. It is worth noting that the variable inside the parentheses \{\} is replaced with its actual values.}
    \label{tab:prompt}
    \vspace{-0.075in}
    \resizebox{1\textwidth}{!}{
        \begin{tabular}{ll}
        \toprule
        \multicolumn{1}{p{.3\textwidth}}{\textbf{Types}} & \makecell{\multicolumn{1}{p{0.8\textwidth}}{\textbf{Prompts}}} \\
        \midrule
        \multicolumn{1}{p{.3\textwidth}}{\textbf{Knowledge Generation}} & 
        \makecell{
            \multicolumn{1}{p{0.8\textwidth}}{DB Schema: \{Database Schema\}} \\
            \noalign{\vskip 0.25ex}\cdashline{1-1}\noalign{\vskip 0.75ex}
            \multicolumn{1}{p{0.8\textwidth}}{Question: \{Few-Shot Question 1\}} \\
            \multicolumn{1}{p{0.8\textwidth}}{Evidence: \{Few-Shot Evidence 1\}} \\
            \noalign{\vskip 0.25ex}\cdashline{1-1}\noalign{\vskip 0.75ex}
            \multicolumn{1}{p{0.8\textwidth}}{Question: \{Few-Shot Question 2\}} \\
            \multicolumn{1}{p{0.8\textwidth}}{Evidence: \{Few-Shot Evidence 2\}} \\
            \noalign{\vskip 0.25ex}\cdashline{1-1}\noalign{\vskip 0.75ex}
            \multicolumn{1}{p{0.8\textwidth}}{...} \\
            \noalign{\vskip 0.25ex}\cdashline{1-1}\noalign{\vskip 0.75ex}
            \multicolumn{1}{p{0.8\textwidth}}{Question: \{Few-Shot Question 10\}} \\
            \multicolumn{1}{p{0.8\textwidth}}{Evidence: \{Few-Shot Evidence 10\}} \\
            \noalign{\vskip 0.25ex}\cdashline{1-1}\noalign{\vskip 0.75ex}
            \multicolumn{1}{p{0.8\textwidth}}{Question: \{Target Question\}} \\
            \multicolumn{1}{p{0.8\textwidth}}{Evidence: } \\
        }\\
        \midrule
        \multicolumn{1}{p{.3\textwidth}}{\textbf{SQL Generation}} & 
        \makecell{
            \multicolumn{1}{p{0.8\textwidth}}{DB Schema: \{Database Schema\}} \\
            \noalign{\vskip 0.25ex}\cdashline{1-1}\noalign{\vskip 0.75ex}
            \multicolumn{1}{p{0.8\textwidth}}{Question: \{Few-Shot Question 1\}} \\
            \multicolumn{1}{p{0.8\textwidth}}{Evidence: \{Few-Shot Evidence 1\}} \\
            \multicolumn{1}{p{0.8\textwidth}}{SQL: \{Few-Shot SQL 1\}} \\
            \noalign{\vskip 0.25ex}\cdashline{1-1}\noalign{\vskip 0.75ex}
            \multicolumn{1}{p{0.8\textwidth}}{Question: \{Few-Shot Question 2\}} \\
            \multicolumn{1}{p{0.8\textwidth}}{Evidence: \{Few-Shot Evidence 2\}} \\
            \multicolumn{1}{p{0.8\textwidth}}{SQL: \{Few-Shot SQL 2\}} \\
            \noalign{\vskip 0.25ex}\cdashline{1-1}\noalign{\vskip 0.75ex}
            \multicolumn{1}{p{0.8\textwidth}}{...} \\
            \noalign{\vskip 0.25ex}\cdashline{1-1}\noalign{\vskip 0.75ex}
            \multicolumn{1}{p{0.8\textwidth}}{Question: \{Few-Shot Question 10\}} \\
            \multicolumn{1}{p{0.8\textwidth}}{Evidence: \{Few-Shot Evidence 10\}} \\
            \multicolumn{1}{p{0.8\textwidth}}{SQL: \{Few-Shot SQL 10\}} \\
            \noalign{\vskip 0.25ex}\cdashline{1-1}\noalign{\vskip 0.75ex}
            \multicolumn{1}{p{0.8\textwidth}}{Question: \{Target Question\}} \\
            \multicolumn{1}{p{0.8\textwidth}}{Evidence: \{Generated Knowledge\}} \\
            \multicolumn{1}{p{0.8\textwidth}}{SQL: } \\
        } \\
        \bottomrule
        \end{tabular}
    }
\end{table*}

%% file: Algorithms/KB-Construction.tex
\begin{algorithm*}[t]
\caption{Knowledge Base Construction for KAT-SQL}
\label{alg:knowledge_base_construction}
\begin{algorithmic}[1]
\Require Dataset $\mD$ containing query-schema-knowledge triplets $(\vq, \mathcal{D}, \vk)$; Prompt template $\mathcal{T}$
\Ensure Knowledge base $\mathcal{K}$
\State $\mathcal{K} \gets \{\}$ \Comment{Initialize an empty knowledge base}
\ForAll{$(\vq, \mathcal{D}, \vk) \in \mD$}
    \State $\mathcal{K} \gets \mathcal{K} \cup \vk$ \Comment{Add existing knowledge to the knowledge base}
\EndFor
\ForAll{query-schema pair $(\vq, \mathcal{D}) \in \mD$}
    \State $\mathcal{E} \gets \text{Top-}k$ relevant examples to the query $\vq$ from $\mD$
    \For{$i = 1 \text{ to } N$} \Comment{Iteratively expand knowledge}
        \State $\mathcal{E}_{\text{perm}} \gets$ Permute examples $\mathcal{E}$
        \State $\vk_{\text{new}} \gets \texttt{LLM}(\mathcal{T}(\vq, \mathcal{D}, \mathcal{E}_{\text{perm}}))$ \Comment{Generate knowledge using LLM with examples}
        \State $\mathcal{K} \gets \mathcal{K} \cup \vk_{\text{new}}$ \Comment{Store generated knowledge in the knowledge base}
    \EndFor
\EndFor
\end{algorithmic}
\end{algorithm*}

%% file: Algorithms/KAT-SQL.tex
\begin{algorithm*}[t]
\caption{Knowledge-Augmented Text-to-SQL (KAT-SQL)}
\label{alg:kat-sql}
\begin{algorithmic}[1]
\Require Query $\vq$; Database schema $\mathcal{D}$; Knowledge base $\mathcal{K}$
\Ensure SQL statement $\vs$
\vspace{0.05in}
\Function{KAT-SQL}{$\vq$, $\mathcal{D}$, $\mathcal{K}$}
    \State $\{\vk_i\}_{i=1}^j \gets$ \Call{Retriever}{$\vq$, $\mathcal{K}$} \Comment{Retrieve relevant knowledge entries from $\mathcal{K}$}
    \State $\mathcal{T}_{\text{ref}} \gets$ \Call{CreatePrompt}{$\vq$, $\{\vk_i\}_{i=1}^j$, $\mathcal{D}$} \Comment{Construct the prompt with retrieved knowledge}
    \State $\vk' \gets \texttt{LLM}(\mathcal{T}_{\text{ref}})$ \Comment{Refine knowledge using LLM}
    \State $\mathcal{T}_{\text{aug}} \gets$ \Call{CreatePrompt}{$\vq$, $\vk'$, $\mathcal{D}$} \Comment{Augment the prompt with refined knowledge}
    \State $\vs \gets \texttt{LLM}(\mathcal{T}_{\text{aug}})$ \Comment{Generate SQL with knowledge augmentation}
    \State \Return $\vs$
\EndFunction
\vspace{0.05in}

\Function{Retriever}{$\vq$, $\mathcal{K}$}
    \State Compute embeddings for $\vq$ and all knowledge entries $\vk \in \mathcal{K}$
    \State Retrieve top-$j$ relevant knowledge entries $\{\vk_i\}_{i=1}^j$ based on embedding similarities
    \State \Return $\{\vk_i\}_{i=1}^j$
\EndFunction
\vspace{0.05in}

\Function{CreatePrompt}{$\vq$, $\vk$, $\mathcal{D}$}
    \State Construct the prompt template $\mathcal{T}$ using the query $\vq$, knowledge $\vk$, and database schema $\mathcal{D}$
    \State \Return $\mathcal{T}$
\EndFunction
\vspace{0.05in}

\end{algorithmic}
\end{algorithm*}

%% file: acl_latex.bbl
\begin{thebibliography}{39}
\providecommand{\natexlab}[1]{#1}

\bibitem[{AI@Meta(2024)}]{llama3modelcard}
AI@Meta. 2024.
\newblock \href {https://github.com/meta-llama/llama3/blob/main/MODEL_CARD.md} {Llama 3 model card}.

\bibitem[{Alivanistos et~al.(2022)Alivanistos, Santamar{\'{\i}}a, Cochez, Kalo, van Krieken, and Thanapalasingam}]{LLM/KBC/1}
Dimitrios Alivanistos, Selene~Baez Santamar{\'{\i}}a, Michael Cochez, Jan{-}Christoph Kalo, Emile van Krieken, and Thiviyan Thanapalasingam. 2022.
\newblock \href {http://ceur-ws.org/Vol-3274/paper2.pdf} {Prompting as probing: Using language models for knowledge base construction}.
\newblock In \emph{Proceedings of the Semantic Web Challenge on Knowledge Base Construction from Pre-trained Language Models 2022 co-located with the 21st International Semantic Web Conference (ISWC2022), Virtual Event, Hanghzou, China, October 2022}, volume 3274 of \emph{{CEUR} Workshop Proceedings}, pages 11--34. CEUR-WS.org.

\bibitem[{Anil et~al.(2023)Anil, Borgeaud, Wu, Alayrac, Yu, Soricut, Schalkwyk, Dai, Hauth, Millican, Silver, Petrov, Johnson, Antonoglou, Schrittwieser, Glaese, Chen, Pitler, Lillicrap, Lazaridou, Firat, Molloy, Isard, Barham, Hennigan, Lee, Viola, Reynolds, Xu, Doherty, Collins, Meyer, Rutherford, Moreira, Ayoub, Goel, Tucker, Piqueras, Krikun, Barr, Savinov, Danihelka, Roelofs, White, Andreassen, von Glehn, Yagati, Kazemi, Gonzalez, Khalman, Sygnowski, and et~al.}]{Gemini}
Rohan Anil, Sebastian Borgeaud, Yonghui Wu, Jean{-}Baptiste Alayrac, Jiahui Yu, Radu Soricut, Johan Schalkwyk, Andrew~M. Dai, Anja Hauth, Katie Millican, David Silver, Slav Petrov, Melvin Johnson, Ioannis Antonoglou, Julian Schrittwieser, Amelia Glaese, Jilin Chen, Emily Pitler, Timothy~P. Lillicrap, Angeliki Lazaridou, Orhan Firat, James Molloy, Michael Isard, Paul~Ronald Barham, Tom Hennigan, Benjamin Lee, Fabio Viola, Malcolm Reynolds, Yuanzhong Xu, Ryan Doherty, Eli Collins, Clemens Meyer, Eliza Rutherford, Erica Moreira, Kareem Ayoub, Megha Goel, George Tucker, Enrique Piqueras, Maxim Krikun, Iain Barr, Nikolay Savinov, Ivo Danihelka, Becca Roelofs, Ana{\"{\i}}s White, Anders Andreassen, Tamara von Glehn, Lakshman Yagati, Mehran Kazemi, Lucas Gonzalez, Misha Khalman, Jakub Sygnowski, and et~al. 2023.
\newblock \href {https://doi.org/10.48550/arXiv.2312.11805} {Gemini: {A} family of highly capable multimodal models}.
\newblock \emph{arXiv preprint arXiv:2312.11805}.

\bibitem[{Cai et~al.(2018)Cai, Xu, Zhang, Yang, Li, and Liang}]{text2sql/4}
Ruichu Cai, Boyan Xu, Zhenjie Zhang, Xiaoyan Yang, Zijian Li, and Zhihao Liang. 2018.
\newblock \href {https://doi.org/10.24963/ijcai.2018/553} {An encoder-decoder framework translating natural language to database queries}.
\newblock In \emph{Proceedings of the Twenty-Seventh International Joint Conference on Artificial Intelligence, {IJCAI} 2018, July 13-19, 2018, Stockholm, Sweden}, pages 3977--3983. ijcai.org.

\bibitem[{Chang and Fosler{-}Lussier(2023)}]{text2sql/prompt/1}
Shuaichen Chang and Eric Fosler{-}Lussier. 2023.
\newblock \href {https://doi.org/10.48550/arXiv.2305.11853} {How to prompt llms for text-to-sql: {A} study in zero-shot, single-domain, and cross-domain settings}.
\newblock \emph{arXiv preprint arXiv:2305.11853}.

\bibitem[{Devlin et~al.(2019)Devlin, Chang, Lee, and Toutanova}]{BERT}
Jacob Devlin, Ming{-}Wei Chang, Kenton Lee, and Kristina Toutanova. 2019.
\newblock \href {https://doi.org/10.18653/V1/N19-1423} {{BERT:} pre-training of deep bidirectional transformers for language understanding}.
\newblock In \emph{Proceedings of the 2019 Conference of the North American Chapter of the Association for Computational Linguistics: Human Language Technologies, {NAACL-HLT} 2019, Minneapolis, MN, USA, June 2-7, 2019, Volume 1 (Long and Short Papers)}, pages 4171--4186. Association for Computational Linguistics.

\bibitem[{Dong et~al.(2023)Dong, Zhang, Ge, Mao, Gao, Chen, Lin, and Lou}]{C3}
Xuemei Dong, Chao Zhang, Yuhang Ge, Yuren Mao, Yunjun Gao, Lu~Chen, Jinshu Lin, and Dongfang Lou. 2023.
\newblock \href {https://doi.org/10.48550/arXiv.2307.07306} {{C3:} zero-shot text-to-sql with chatgpt}.
\newblock \emph{arXiv preprint arXiv:2307.07306}.

\bibitem[{Dou et~al.(2022)Dou, Gao, Liu, Pan, Wang, Che, Zhan, Kan, and Lou}]{knowledge-intensive-text-to-sql}
Longxu Dou, Yan Gao, Xuqi Liu, Mingyang Pan, Dingzirui Wang, Wanxiang Che, Dechen Zhan, Min{-}Yen Kan, and Jian{-}Guang Lou. 2022.
\newblock \href {https://doi.org/10.18653/v1/2022.emnlp-main.350} {Towards knowledge-intensive text-to-sql semantic parsing with formulaic knowledge}.
\newblock In \emph{Proceedings of the 2022 Conference on Empirical Methods in Natural Language Processing, {EMNLP} 2022, Abu Dhabi, United Arab Emirates, December 7-11, 2022}, pages 5240--5253. Association for Computational Linguistics.

\bibitem[{Douze et~al.(2024)Douze, Guzhva, Deng, Johnson, Szilvasy, Mazar{\'{e}}, Lomeli, Hosseini, and J{\'{e}}gou}]{faiss}
Matthijs Douze, Alexandr Guzhva, Chengqi Deng, Jeff Johnson, Gergely Szilvasy, Pierre{-}Emmanuel Mazar{\'{e}}, Maria Lomeli, Lucas Hosseini, and Herv{\'{e}} J{\'{e}}gou. 2024.
\newblock \href {https://doi.org/10.48550/ARXIV.2401.08281} {The faiss library}.
\newblock \emph{arXiv preprint arXiv:2401.08281}.

\bibitem[{Gao et~al.(2024)Gao, Wang, Li, Sun, Qian, Ding, and Zhou}]{LLM/Text2SQL/2}
Dawei Gao, Haibin Wang, Yaliang Li, Xiuyu Sun, Yichen Qian, Bolin Ding, and Jingren Zhou. 2024.
\newblock \href {https://www.vldb.org/pvldb/vol17/p1132-gao.pdf} {Text-to-sql empowered by large language models: {A} benchmark evaluation}.
\newblock \emph{Proc. {VLDB} Endow.}, 17(5):1132--1145.

\bibitem[{Gu et~al.(2023)Gu, Fan, Tang, Cao, Jia, Madden, and Du}]{SC-Promp}
Zihui Gu, Ju~Fan, Nan Tang, Lei Cao, Bowen Jia, Sam Madden, and Xiaoyong Du. 2023.
\newblock \href {https://doi.org/10.1145/3589292} {Few-shot text-to-sql translation using structure and content prompt learning}.
\newblock \emph{Proc. {ACM} Manag. Data}, 1(2):147:1--147:28.

\bibitem[{Hofst{\"{a}}tter et~al.(2021)Hofst{\"{a}}tter, Lin, Yang, Lin, and Hanbury}]{TAS-B}
Sebastian Hofst{\"{a}}tter, Sheng{-}Chieh Lin, Jheng{-}Hong Yang, Jimmy Lin, and Allan Hanbury. 2021.
\newblock \href {https://doi.org/10.1145/3404835.3462891} {Efficiently teaching an effective dense retriever with balanced topic aware sampling}.
\newblock In \emph{{SIGIR} '21: The 44th International {ACM} {SIGIR} Conference on Research and Development in Information Retrieval, Virtual Event, Canada, July 11-15, 2021}, pages 113--122. {ACM}.

\bibitem[{Hong et~al.(2024)Hong, Yuan, Chen, Zhang, Huang, and Huang}]{Knowledge-to-SQL}
Zijin Hong, Zheng Yuan, Hao Chen, Qinggang Zhang, Feiran Huang, and Xiao Huang. 2024.
\newblock \href {https://doi.org/10.48550/arXiv.2402.11517} {Knowledge-to-sql: Enhancing {SQL} generation with data expert {LLM}}.
\newblock \emph{arXiv preprint arXiv:2402.11517}.

\bibitem[{Jiang et~al.(2024)Jiang, Sablayrolles, Roux, Mensch, Savary, Bamford, Chaplot, de~Las~Casas, Hanna, Bressand, Lengyel, Bour, Lample, Lavaud, Saulnier, Lachaux, Stock, Subramanian, Yang, Antoniak, Scao, Gervet, Lavril, Wang, Lacroix, and Sayed}]{Mixtral}
Albert~Q. Jiang, Alexandre Sablayrolles, Antoine Roux, Arthur Mensch, Blanche Savary, Chris Bamford, Devendra~Singh Chaplot, Diego de~Las~Casas, Emma~Bou Hanna, Florian Bressand, Gianna Lengyel, Guillaume Bour, Guillaume Lample, L{\'{e}}lio~Renard Lavaud, Lucile Saulnier, Marie{-}Anne Lachaux, Pierre Stock, Sandeep Subramanian, Sophia Yang, Szymon Antoniak, Teven~Le Scao, Th{\'{e}}ophile Gervet, Thibaut Lavril, Thomas Wang, Timoth{\'{e}}e Lacroix, and William~El Sayed. 2024.
\newblock \href {https://doi.org/10.48550/arXiv.2401.04088} {Mixtral of experts}.
\newblock \emph{arXiv preprint arXiv:2401.04088}.

\bibitem[{Karpukhin et~al.(2020)Karpukhin, Oguz, Min, Lewis, Wu, Edunov, Chen, and Yih}]{dpr}
Vladimir Karpukhin, Barlas Oguz, Sewon Min, Patrick S.~H. Lewis, Ledell Wu, Sergey Edunov, Danqi Chen, and Wen{-}tau Yih. 2020.
\newblock \href {https://doi.org/10.18653/v1/2020.emnlp-main.550} {Dense passage retrieval for open-domain question answering}.
\newblock In \emph{Proceedings of the 2020 Conference on Empirical Methods in Natural Language Processing, {EMNLP} 2020, Online, November 16-20, 2020}, pages 6769--6781. Association for Computational Linguistics.

\bibitem[{Lee et~al.(2024)Lee, Park, Kim, and Park}]{text2sql/consistent/1}
Dongjun Lee, Choongwon Park, Jaehyuk Kim, and Heesoo Park. 2024.
\newblock \href {https://doi.org/10.48550/arXiv.2405.07467} {{MCS-SQL:} leveraging multiple prompts and multiple-choice selection for text-to-sql generation}.
\newblock \emph{arXiv preprint arXiv:2405.07467}.

\bibitem[{Li et~al.(2023)Li, Hui, Qu, Yang, Li, Li, Wang, Qin, Geng, Huo, Zhou, Ma, Li, Chang, Huang, Cheng, and Li}]{bird}
Jinyang Li, Binyuan Hui, Ge~Qu, Jiaxi Yang, Binhua Li, Bowen Li, Bailin Wang, Bowen Qin, Ruiying Geng, Nan Huo, Xuanhe Zhou, Chenhao Ma, Guoliang Li, Kevin~Chen{-}Chuan Chang, Fei Huang, Reynold Cheng, and Yongbin Li. 2023.
\newblock \href {http://papers.nips.cc/paper\_files/paper/2023/hash/83fc8fab1710363050bbd1d4b8cc0021-Abstract-Datasets\_and\_Benchmarks.html} {Can {LLM} already serve as {A} database interface? {A} big bench for large-scale database grounded text-to-sqls}.
\newblock In \emph{Advances in Neural Information Processing Systems 36: Annual Conference on Neural Information Processing Systems 2023, NeurIPS 2023, New Orleans, LA, USA, December 10 - 16, 2023}.

\bibitem[{Liu and Tan(2023)}]{text2sql/cot/1}
Xiping Liu and Zhao Tan. 2023.
\newblock \href {https://doi.org/10.48550/arXiv.2304.11556} {Divide and prompt: Chain of thought prompting for text-to-sql}.
\newblock \emph{arXiv preprint arXiv:2304.11556}.

\bibitem[{Mishra et~al.(2024)Mishra, Stallone, Zhang, Shen, Prasad, Soria, Merler, Selvam, Surendran, Singh, Sethi, Dang, Li, Wu, Zawad, Coleman, White, Lewis, Pavuluri, Koyfman, Lublinsky, de~Bayser, Abdelaziz, Basu, Agarwal, Zhou, Johnson, Goyal, Patel, Shah, Zerfos, Ludwig, Munawar, Crouse, Kapanipathi, Salaria, Calio, Wen, Seelam, Belgodere, Fonseca, Singhee, Desai, Cox, Puri, and Panda}]{Granite}
Mayank Mishra, Matt Stallone, Gaoyuan Zhang, Yikang Shen, Aditya Prasad, Adriana~Meza Soria, Michele Merler, Parameswaran Selvam, Saptha Surendran, Shivdeep Singh, Manish Sethi, Xuan{-}Hong Dang, Pengyuan Li, Kun{-}Lung Wu, Syed Zawad, Andrew Coleman, Matthew White, Mark Lewis, Raju Pavuluri, Yan Koyfman, Boris Lublinsky, Maximilien de~Bayser, Ibrahim Abdelaziz, Kinjal Basu, Mayank Agarwal, Yi~Zhou, Chris Johnson, Aanchal Goyal, Hima Patel, S.~Yousaf Shah, Petros Zerfos, Heiko Ludwig, Asim Munawar, Maxwell Crouse, Pavan Kapanipathi, Shweta Salaria, Bob Calio, Sophia Wen, Seetharami Seelam, Brian Belgodere, Carlos~A. Fonseca, Amith Singhee, Nirmit Desai, David~D. Cox, Ruchir Puri, and Rameswar Panda. 2024.
\newblock \href {https://doi.org/10.48550/arXiv.2405.04324} {Granite code models: {A} family of open foundation models for code intelligence}.
\newblock \emph{arXiv preprint arXiv:2405.04324}.

\bibitem[{Mukherjee et~al.(2023)Mukherjee, Mitra, Jawahar, Agarwal, Palangi, and Awadallah}]{orca}
Subhabrata Mukherjee, Arindam Mitra, Ganesh Jawahar, Sahaj Agarwal, Hamid Palangi, and Ahmed Awadallah. 2023.
\newblock \href {https://doi.org/10.48550/arXiv.2306.02707} {Orca: Progressive learning from complex explanation traces of {GPT-4}}.
\newblock \emph{arXiv preprint arXiv:2306.02707}.

\bibitem[{Nayak and Timmapathini(2023)}]{LLM/KBC/2}
Anmol Nayak and Hariprasad Timmapathini. 2023.
\newblock \href {https://ceur-ws.org/Vol-3577/paper7.pdf} {{LLM2KB:} constructing knowledge bases using instruction tuned context aware large language models}.
\newblock In \emph{Joint proceedings of the 1st workshop on Knowledge Base Construction from Pre-Trained Language Models {(KBC-LM)} and the 2nd challenge on Language Models for Knowledge Base Construction {(LM-KBC)} co-located with the 22nd International Semantic Web Conference {(ISWC} 2023), Athens, Greece, November 6, 2023}, volume 3577 of \emph{{CEUR} Workshop Proceedings}. CEUR-WS.org.

\bibitem[{OpenAI(2023)}]{GPT-4}
OpenAI. 2023.
\newblock \href {https://doi.org/10.48550/ARXIV.2303.08774} {{GPT-4} technical report}.
\newblock \emph{arXiv preprint arXiv:2303.08774}.

\bibitem[{Pourreza and Rafiei(2023)}]{DIN-SQL}
Mohammadreza Pourreza and Davood Rafiei. 2023.
\newblock \href {http://papers.nips.cc/paper\_files/paper/2023/hash/72223cc66f63ca1aa59edaec1b3670e6-Abstract-Conference.html} {{DIN-SQL:} decomposed in-context learning of text-to-sql with self-correction}.
\newblock In \emph{Advances in Neural Information Processing Systems 36: Annual Conference on Neural Information Processing Systems 2023, NeurIPS 2023, New Orleans, LA, USA, December 10 - 16, 2023}.

\bibitem[{Rajkumar et~al.(2022)Rajkumar, Li, and Bahdanau}]{LLM/Text2SQL/1}
Nitarshan Rajkumar, Raymond Li, and Dzmitry Bahdanau. 2022.
\newblock \href {https://doi.org/10.48550/arXiv.2204.00498} {Evaluating the text-to-sql capabilities of large language models}.
\newblock \emph{arXiv preprint arXiv:2204.00498}.

\bibitem[{Song et~al.(2020)Song, Tan, Qin, Lu, and Liu}]{MPNet}
Kaitao Song, Xu~Tan, Tao Qin, Jianfeng Lu, and Tie{-}Yan Liu. 2020.
\newblock \href {https://proceedings.neurips.cc/paper/2020/hash/c3a690be93aa602ee2dc0ccab5b7b67e-Abstract.html} {Mpnet: Masked and permuted pre-training for language understanding}.
\newblock In \emph{Advances in Neural Information Processing Systems 33: Annual Conference on Neural Information Processing Systems 2020, NeurIPS 2020, December 6-12, 2020, virtual}.

\bibitem[{Sudalairaj et~al.(2024)Sudalairaj, Bhandwaldar, Pareja, Xu, Cox, and Srivastava}]{instructlab}
Shivchander Sudalairaj, Abhishek Bhandwaldar, Aldo Pareja, Kai Xu, David~D. Cox, and Akash Srivastava. 2024.
\newblock \href {https://doi.org/10.48550/arXiv.2403.01081} {{LAB:} large-scale alignment for chatbots}.
\newblock \emph{arXiv preprint arXiv:2403.01081}.

\bibitem[{Tai et~al.(2023)Tai, Chen, Zhang, Deng, and Sun}]{text2sql/cot/2}
Chang{-}Yu Tai, Ziru Chen, Tianshu Zhang, Xiang Deng, and Huan Sun. 2023.
\newblock \href {https://doi.org/10.18653/v1/2023.emnlp-main.327} {Exploring chain of thought style prompting for text-to-sql}.
\newblock In \emph{Proceedings of the 2023 Conference on Empirical Methods in Natural Language Processing, {EMNLP} 2023, Singapore, December 6-10, 2023}, pages 5376--5393. Association for Computational Linguistics.

\bibitem[{Taori et~al.(2023)Taori, Gulrajani, Zhang, Dubois, Li, Guestrin, Liang, and Hashimoto}]{alpaca}
Rohan Taori, Ishaan Gulrajani, Tianyi Zhang, Yann Dubois, Xuechen Li, Carlos Guestrin, Percy Liang, and Tatsunori~B. Hashimoto. 2023.
\newblock Stanford alpaca: An instruction-following llama model.
\newblock \url{https://github.com/tatsu-lab/stanford_alpaca}.

\bibitem[{Veseli et~al.(2023)Veseli, Razniewski, Kalo, and Weikum}]{LLM/KBC/3}
Blerta Veseli, Simon Razniewski, Jan{-}Christoph Kalo, and Gerhard Weikum. 2023.
\newblock \href {https://doi.org/10.18653/v1/2023.findings-emnlp.426} {Evaluating the knowledge base completion potential of {GPT}}.
\newblock In \emph{Findings of the Association for Computational Linguistics: {EMNLP} 2023, Singapore, December 6-10, 2023}, pages 6432--6443. Association for Computational Linguistics.

\bibitem[{Vrandecic and Kr{\"{o}}tzsch(2014)}]{Wikidata}
Denny Vrandecic and Markus Kr{\"{o}}tzsch. 2014.
\newblock \href {https://doi.org/10.1145/2629489} {Wikidata: a free collaborative knowledgebase}.
\newblock \emph{Commun. {ACM}}, 57(10):78--85.

\bibitem[{Wang et~al.(2023{\natexlab{a}})Wang, Ren, Yang, Liang, Bai, Zhang, Yan, and Li}]{MAC-SQL}
Bing Wang, Changyu Ren, Jian Yang, Xinnian Liang, Jiaqi Bai, Qian{-}Wen Zhang, Zhao Yan, and Zhoujun Li. 2023{\natexlab{a}}.
\newblock \href {https://doi.org/10.48550/arXiv.2312.11242} {{MAC-SQL:} {A} multi-agent collaborative framework for text-to-sql}.
\newblock \emph{arXiv preprint arXiv:2312.11242}.

\bibitem[{Wang et~al.(2023{\natexlab{b}})Wang, Wei, Schuurmans, Le, Chi, Narang, Chowdhery, and Zhou}]{Self-Consistency}
Xuezhi Wang, Jason Wei, Dale Schuurmans, Quoc~V. Le, Ed~H. Chi, Sharan Narang, Aakanksha Chowdhery, and Denny Zhou. 2023{\natexlab{b}}.
\newblock \href {https://openreview.net/pdf?id=1PL1NIMMrw} {Self-consistency improves chain of thought reasoning in language models}.
\newblock In \emph{The Eleventh International Conference on Learning Representations, {ICLR} 2023, Kigali, Rwanda, May 1-5, 2023}. OpenReview.net.

\bibitem[{Wang et~al.(2023{\natexlab{c}})Wang, Kordi, Mishra, Liu, Smith, Khashabi, and Hajishirzi}]{Self-Instruct}
Yizhong Wang, Yeganeh Kordi, Swaroop Mishra, Alisa Liu, Noah~A. Smith, Daniel Khashabi, and Hannaneh Hajishirzi. 2023{\natexlab{c}}.
\newblock \href {https://doi.org/10.18653/v1/2023.acl-long.754} {Self-instruct: Aligning language models with self-generated instructions}.
\newblock In \emph{Proceedings of the 61st Annual Meeting of the Association for Computational Linguistics (Volume 1: Long Papers), {ACL} 2023, Toronto, Canada, July 9-14, 2023}, pages 13484--13508. Association for Computational Linguistics.

\bibitem[{Wei et~al.(2022)Wei, Wang, Schuurmans, Bosma, Ichter, Xia, Chi, Le, and Zhou}]{CoT}
Jason Wei, Xuezhi Wang, Dale Schuurmans, Maarten Bosma, Brian Ichter, Fei Xia, Ed~H. Chi, Quoc~V. Le, and Denny Zhou. 2022.
\newblock \href {http://papers.nips.cc/paper\_files/paper/2022/hash/9d5609613524ecf4f15af0f7b31abca4-Abstract-Conference.html} {Chain-of-thought prompting elicits reasoning in large language models}.
\newblock In \emph{Advances in Neural Information Processing Systems 35: Annual Conference on Neural Information Processing Systems 2022, NeurIPS 2022, New Orleans, LA, USA, November 28 - December 9, 2022}.

\bibitem[{Xu et~al.(2023)Xu, Sun, Zheng, Geng, Zhao, Feng, Tao, and Jiang}]{evol-instruct}
Can Xu, Qingfeng Sun, Kai Zheng, Xiubo Geng, Pu~Zhao, Jiazhan Feng, Chongyang Tao, and Daxin Jiang. 2023.
\newblock \href {https://doi.org/10.48550/arXiv.2304.12244} {Wizardlm: Empowering large language models to follow complex instructions}.
\newblock \emph{arXiv preprint arXiv:2304.12244}.

\bibitem[{Xu et~al.(2017)Xu, Liu, and Song}]{text2sql/2}
Xiaojun Xu, Chang Liu, and Dawn Song. 2017.
\newblock \href {http://arxiv.org/abs/1711.04436} {Sqlnet: Generating structured queries from natural language without reinforcement learning}.
\newblock \emph{arXiv preprint arXiv:1711.04436}.

\bibitem[{Yaghmazadeh et~al.(2017)Yaghmazadeh, Wang, Dillig, and Dillig}]{text2sql/3}
Navid Yaghmazadeh, Yuepeng Wang, Isil Dillig, and Thomas Dillig. 2017.
\newblock \href {https://doi.org/10.1145/3133887} {Sqlizer: query synthesis from natural language}.
\newblock \emph{Proc. {ACM} Program. Lang.}, 1({OOPSLA}):63:1--63:26.

\bibitem[{Yu et~al.(2018)Yu, Zhang, Yang, Yasunaga, Wang, Li, Ma, Li, Yao, Roman, Zhang, and Radev}]{Spider}
Tao Yu, Rui Zhang, Kai Yang, Michihiro Yasunaga, Dongxu Wang, Zifan Li, James Ma, Irene Li, Qingning Yao, Shanelle Roman, Zilin Zhang, and Dragomir~R. Radev. 2018.
\newblock \href {https://doi.org/10.18653/v1/d18-1425} {Spider: {A} large-scale human-labeled dataset for complex and cross-domain semantic parsing and text-to-sql task}.
\newblock In \emph{Proceedings of the 2018 Conference on Empirical Methods in Natural Language Processing, Brussels, Belgium, October 31 - November 4, 2018}, pages 3911--3921. Association for Computational Linguistics.

\bibitem[{Zelle and Mooney(1996)}]{text2sql/1}
John~M. Zelle and Raymond~J. Mooney. 1996.
\newblock \href {http://www.aaai.org/Library/AAAI/1996/aaai96-156.php} {Learning to parse database queries using inductive logic programming}.
\newblock In \emph{Proceedings of the Thirteenth National Conference on Artificial Intelligence and Eighth Innovative Applications of Artificial Intelligence Conference, {AAAI} 96, {IAAI} 96, Portland, Oregon, USA, August 4-8, 1996, Volume 2}, pages 1050--1055. {AAAI} Press / The {MIT} Press.

\end{thebibliography}
